\documentclass[conference]{IEEEtran}
\IEEEoverridecommandlockouts
\usepackage{cite}
\usepackage{amsmath,amssymb,amsfonts}
\usepackage{algorithm}
\usepackage{algpseudocode}
\usepackage{graphicx}
\usepackage{url}
\usepackage{hyperref}
\usepackage{textcomp}
\usepackage{booktabs}
\usepackage{xcolor}
\DeclareMathAlphabet{\mathpzc}{OT1}{pzc}{m}{it}
\usepackage{placeins}
\usepackage{multirow}
\usepackage{subcaption}
\usepackage{breqn}
\usepackage{enumitem}
\usepackage{array}
\usepackage{tabularx}
\usepackage{amsthm}
\def\BibTeX{{\rm B\kern-.05em{\sc i\kern-.025em b}\kern-.08em
    T\kern-.1667em\lower.7ex\hbox{E}\kern-.125emX}}
\usepackage{color, colortbl}	
\definecolor{LightCyan}{rgb}{0.88,1,1}
\definecolor{Gray}{gray}{0.90}

\begin{document}

\title{Optimizing Privacy and Utility Tradeoffs for Group Interests Through Harmonization
}
\author{\IEEEauthorblockN{\textsuperscript{}Bishwas Mandal}
\IEEEauthorblockA{\textit{Department of Computer Science} \\
\textit{Kansas State University}\\
 Manhattan, KS 66506, USA \\
 bishmdl76@ksu.edu}
\and
\IEEEauthorblockN{\textsuperscript{}George Amariucai}
 \IEEEauthorblockA{\textit{Department of Computer Science} \\
\textit{Kansas State University}\\
 Manhattan, KS 66506, USA \\
 amariucai@ksu.edu}
\and
\IEEEauthorblockN{\textsuperscript{}Shuangqing Wei}
\IEEEauthorblockA{\textit{Division of Electrical \& Computer Engineering} \\
\textit{Louisiana State University}\\
 Baton Rouge, LA 70803, USA \\
 swei@lsu.edu}
}

\maketitle
\begin{abstract}
We propose a novel problem formulation to address the privacy-utility tradeoff, specifically when dealing with two distinct user groups characterized by unique sets of private and utility attributes. Unlike previous studies that primarily focus on scenarios where all users share identical private and utility attributes and often rely on auxiliary datasets or manual annotations, we introduce a collaborative data-sharing mechanism between two user groups through a trusted third party. This third party uses adversarial privacy techniques with our proposed data-sharing mechanism to internally sanitize data for both groups and eliminates the need for manual annotation or auxiliary datasets. Our methodology ensures that private attributes cannot be accurately inferred while enabling highly accurate predictions of utility features. Importantly, even if analysts or adversaries possess auxiliary datasets containing raw data, they are unable to accurately deduce private features. Additionally, our data-sharing mechanism is compatible with various existing adversarially trained privacy techniques. We empirically demonstrate the effectiveness of our approach using synthetic and real-world datasets, showcasing its ability to balance the conflicting goals of privacy and utility.

\end{abstract}

\begin{IEEEkeywords}
privacy-utility tradeoff, group settings, data publishing, inference privacy, adversarial optimization.
\end{IEEEkeywords}

\section{Introduction}

 \begin{figure*}[!htb]
\centering
  \includegraphics[width=0.97\textwidth]{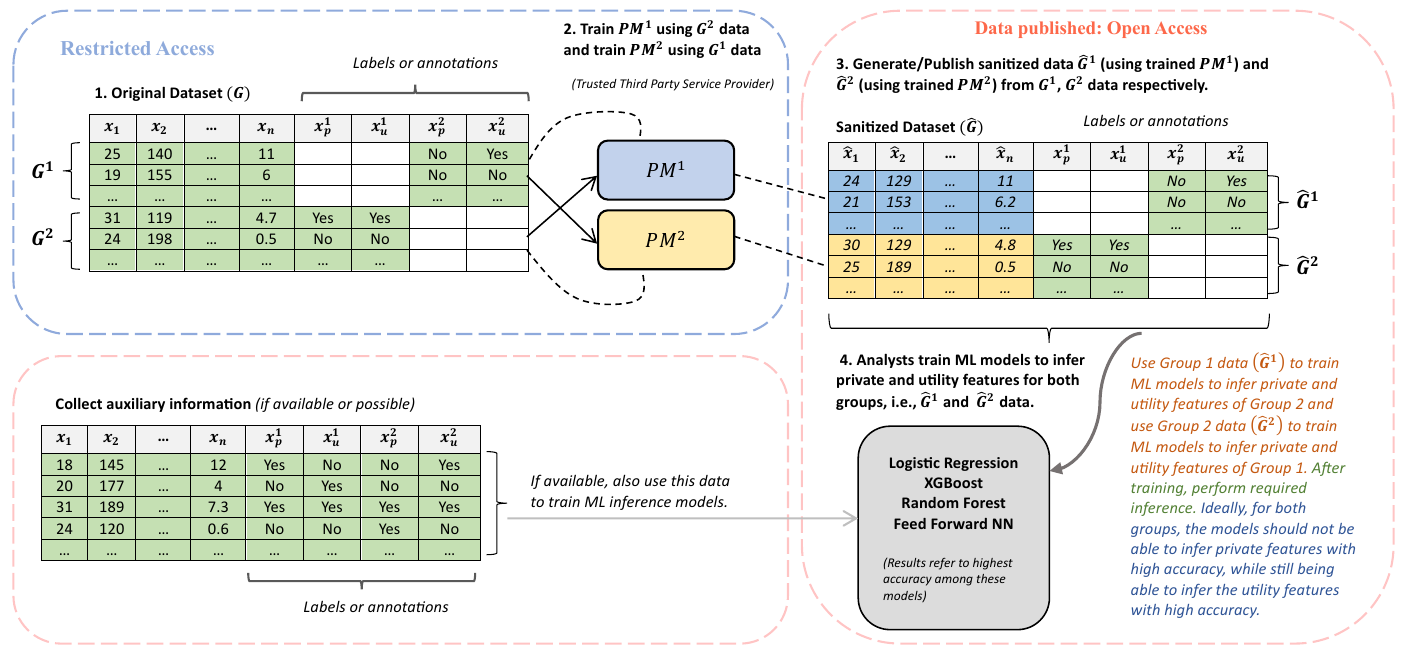}
  \caption{\textmd{Overview of the privacy-utility tradeoff in two-user group setting. \textit{Restricted Access} block demonstrates how data from one group trains the privacy mechanism for another and vice-versa. \textit{Open Access} block details the public release of sanitized data available for public analysis. Blank spaces represent private and utility features of a particular group which are not present in the dataset, and analysts or adversaries aim to make correct predictions of these features.}}
  \label{problem_formulation_setup}
\end{figure*}

Machine learning models have proven to be extremely effective in addressing a diverse range of practical challenges, such as image classification, regression tasks, forecasting human behavior, and processing natural languages, among others. Their capacity to process and make sense of substantial data volumes render them exceedingly valuable. The remarkable achievements of machine learning models into multiple domains have led data analysts and scientists to extensively integrate machine learning in various predictive and generative modeling tasks. While these professionals can utilize the data to better understand user needs and enhance the accuracy of their predictive models, there is also a potential risk of misusing the data to deduce sensitive information about individuals.

In numerous practical applications, tabular data is routinely generated as a standard output of relational database systems \cite{shwartzziv2021tabular}. This research primarily deals with situations where data analysts are tasked with closely examining tabular datasets to create predictive models for various classification tasks. To build such predictive models, analysts have two methods: they can either annotate the data themselves or seek and collect an auxiliary dataset that shares the same input features and contains the labels they aim to predict. In addition to predicting useful features, analysts also have the potential to build models to infer other private or sensitive features using similar methods. Therefore, it is evident that machine learning models, while serving beneficial purposes, can also be used with malicious intent to extract sensitive or private information.

To mitigate the associated privacy risk, existing literature suggests various privacy mechanisms \cite{edwards2016censoring, e19120656, madras2018learning, pittaluga2018learning, 8515092, huang2019generative, chen2019distributed, wu2020privacypreserving, morales2020sensitivenets, Xiao_Tsai_Sohn_Chandraker_Yang_2020, erdemir2021active, wu2021privacypreserving, 9892789} designed to protect sensitive information while still enabling the prediction of useful feature. These mechanisms frequently use adversarial optimization techniques to train their generative models for the data sanitization process. The adversarial optimization methods can be trained either using the auxiliary dataset, if available, or by initially annotating the private and utility labels for the given input data and then training the privacy mechanism. The overarching goal is to develop a privacy mechanism that minimizes the mutual information between the input data and sensitive features while maximizing the mutual information with utility features to safeguard privacy and still enable accurate inferences of utility features.

Previous studies have emphasized privacy (\textit{inference privacy}) concerns associated with analyzing datasets that assume all users share the same private and utility characteristics. However, in real-world situations, it is unrealistic to expect uniform privacy requirements among all individuals within a dataset \cite{4160223}. Consequently, it is likely that diverse user populations, such as those on the internet, possess varying private and utility attributes. These distinct user groups, each with their unique sets of private and utility attributes, are closely interconnected, enabling the transfer of information from one group to another. For instance, collaborative recommendation systems may suggest new items to one group of customers based on the purchase history of another group of customers \cite{1423975}. While this collaborative use of data can have beneficial outcomes, it can also inadvertently reveal sensitive information. For instance, \cite{5696719} demonstrated that utilizing information from an individual's social network connections can effectively disclose details like their location, sexual orientation, marital status, and political affiliation.

In this paper, we introduce a novel problem formulation that revolves around a tabular dataset comprising two distinct user groups, each characterized by unique private and utility features. The primary objective is to achieve highly accurate predictions of the utility features for both user groups while simultaneously ensuring that accurate predictions of their private features are highly improbable. While it is acknowledged that there will be cases where users' private attributes could be correctly inferred, our aim is to keep the accuracy of such predictions intentionally low, thus granting users plausible deniability. In contrast to prior approaches, where tabular datasets typically featured a single set of private and utility features, and the privacy mechanisms being trained relied on auxiliary datasets or manual annotation of data points, our approach has the flexibility to accommodate multiple user groups with varying private and utility features. While our current focus centers on a scenario involving two user groups, our future research endeavors will extend to encompass situations involving multiple user groups. Importantly, our privacy mechanism's training does not depend on manual annotation or auxiliary datasets; it solely relies on the data originating from the two user groups.


\section{Related Work}\label{relatedwork}

In the realm of privacy, there exist two primary categories \cite{8259344, 8766142}. The first category, known as \textit{data privacy}, is primarily focused on safeguarding the original, unprocessed data. In contrast, \textit{inference privacy} seeks to shield sensitive information that can be deduced or inferred from the disclosed raw data, often as a result of correlations. \textit{Raw data}, in this context, refers to the initial, unaltered data collected or generated, representing data in its fundamental state prior to any form of examination, modification, or inference.

Classical methods to preserve data privacy encompass Differential Privacy (DP) \cite{10.1561/0400000042}, and Homomorphic Encryption (HE) \cite{10.1145/3214303}. DP primarily ensures that databases with a single differing record remain indistinguishable, primarily serving as a defense against membership inference attacks. On the other hand, HE allows computations to be performed on encrypted data. It is important to note that these methods do not directly address the issue of inference privacy, which constitutes the central focus of our research. Label Differential Privacy \cite{https://doi.org/10.48550/arxiv.2202.12968} introduces noise to labels to safeguard them, but it does not apply noise to raw data to prevent the inadvertent disclosure of private labels, as is the specific concern within inference privacy.

In a broader context, while differential privacy (DP) doesn't inherently target the challenge of inference privacy, specific adjustments can enable DP to provide a formal guarantee of inference privacy under certain assumptions \cite{ghosh2017inferential}. However, determining the optimal level of noise to introduce into higher-dimensional data for achieving differential privacy can be impractical. Homomorphic Encryption (HE) can be used for inference privacy with certain assumptions, but conducting computations on encrypted data becomes intricate, especially with complex neural network calculations \cite{chen2022thex}. Therefore, recent research on inference privacy has utilized adversarial optimization techniques to minimize the mutual information between disclosed data and private features in various contexts \cite{edwards2016censoring, e19120656, madras2018learning, pittaluga2018learning, 8515092, huang2019generative, chen2019distributed, wu2020privacypreserving, morales2020sensitivenets, Xiao_Tsai_Sohn_Chandraker_Yang_2020, erdemir2021active, wu2021privacypreserving, 9892789}. These studies have primarily focused on a single user group i.e., all users with identical private and utility attributes.


Within the field of data privacy, substantial investigation has been undertaken regarding strategies pertaining to multi-group or multi-user privacy. Various methods involving secure multi-party computation (SMPC) \cite{knott2022crypten} are utilized to preserve data privacy in scenarios featuring multiple parties, ensuring that individual parties are unable to access the data of others. Heterogeneous Differential Privacy \cite{https://doi.org/10.48550/arxiv.1504.06998} takes into account the varying privacy expectations of different users, while Personal Differential Privacy \cite{10.1145/2775051.2677005} allows for individualized privacy budgets for each user, albeit with a trade-off of reduced data utility. To address this challenge,  \cite{9488825} introduced a utility-aware sampling method that caters to the diverse privacy requirements of different users. Another approach, as outlined in \cite{7798524}, presents a model wherein multiple data owners with private real-valued data are shielded using a Gaussian mechanism against multiple data users seeking to glean answers to linear queries. Furthermore, \cite{7862797} provides a differentially private response that ensures the extent of private data leakage depends on the proximity of two users within the network. A comprehensive multi-user privacy approach is presented in \cite{DBLP:journals/popets/GhaziKKMPSWW22}, which combines counting Bloom filters \cite{10.1145/362686.362692}, Differential Privacy (DP), homomorphic encryption (HE), and secure multi-party computation (SMPC) \cite{10.1561/3300000019} to address cardinality and frequency histogram problems.

Despite the considerable body of work dedicated to the exploration of multi-user privacy within the context of data privacy, it is worth noting that, to the best of our knowledge, there exists a noticeable gap in research concerning multi-user inference privacy, particularly when dealing with distinct sets of private and utility features. An investigation presented in \cite{9167255} delves into privacy-utility tradeoffs within multi-agent systems, wherein each agent employs linear compression and noise perturbation techniques to safeguard private information while still facilitating the accurate identification of utility features. In this system, multiple agents operate independently, assuming a scenario devoid of a central trusted entity. However, it's important to distinguish this concept of \textit{multi-agent} from our setup, as in their case, all agents possess uniform private and utility features, whereas we specifically consider varying private and utility features across different user groups. In the subsequent section, we furnish a comprehensive definition of our privacy-utility tradeoff problem involving two user groups.

\section{Problem Formulation}\label{problemformulation}
Consider a scenario where a dataset $G$ comprises $n$ features, denoted as $\mathcal{X} = \{x_1, x_2, x_3, \cdots, x_n\}$. In addition, the dataset $G$ includes labels or annotations represented as $\{x^1_p, x^1_u, x^2_p, x^2_u\}$, where $x^1_p \notin \mathcal{X}, x^1_u \notin \mathcal{X}, x^2_p \notin \mathcal{X}, x^2_u \notin \mathcal{X}$. These features or labels may encompass demographic variables such as gender and race, socio-economic metrics like education level and earnings, or digital interactions like mouse movements, link clicks, and content preferences, depending on the specific application domain. Within this dataset, we assume two distinct user groups: Group 1 (G1) and Group 2 (G2), each consisting of $k/2$ rows, resulting in a total of $k$ rows in the entire dataset $G$. Specifically, G1 includes features and labels denoted as $\mathcal{A} = \mathcal{X} \cup \{x^2_p, x^2_u\}$ and does not contain $\{x^1_p, x^1_u\}$. Here, $x^1_p$ is G1's private feature, intentionally withheld from disclosure, while $x^1_u$ represents the label that G1 aims to predict with high accuracy. Similarly, G2 comprises features and labels denoted as $\mathcal{B} = \mathcal{X} \cup \{x^1_p, x^1_u\}$ and does not include $\{x^2_p, x^2_u\}$. In this context, $x^2_p$ serves as G2's private feature, deliberately omitted from the dataset, while $x^2_u$ signifies the label that G2 seeks to predict accurately. It is important to note that these private and utility features correspond to certain classification labels. For simplicity, we assume binary classification labels in this work. The arrangement of these features is visually depicted in Fig. \ref{problem_formulation_setup}, within the \textit{Restricted Access} block.

If the dataset $G$ were to remain unaltered and be publicly accessible, analysts would possess the capability not only to deduce the valuable utility features ($x^1_u$ for Group 1 and $x^2_u$ for Group 2) but also to discern the sensitive or private features ($x^1_p$ for Group 1 and $x^2_p$ for Group 2). The attainment of this inference is feasible through the application of machine learning (ML) models. As an illustration, to distinguish the private and utility features of G1, ML models can be trained using G2 data, as G2 data contains private and utility labels specific to G1. Similarly, it is possible to achieve a high-precision prediction of the private and utility features of G2. Therefore, the critical necessity lies in the development of a privacy mechanism that can sanitize the dataset such that it allows accurate inference of utility features for both user groups while constraining the accurate inference of private attributes. While recognizing that there may be instances where some users' private attributes can be correctly inferred, our primary aim is to deliberately limit the accuracy of such predictions to provide users with a degree of plausible deniability. In our paper's context, \textit{privacy refers to the adversary’s inability to construct a classifier that can predict the private features with a high accuracy, whereas utility is defined as the ability to construct a classifier that can predict utility features with high accuracy.}

\subsection{Assumptions and Threat Model}\label{assumptions_threat_model}

In our problem formulation, we posit a fundamental absence of trust between Group 1 and Group 2, resulting in a reluctance to engage in direct data sharing for the training of their respective privacy mechanisms (the details of which will be discussed in Section \ref{data-sharing-mechanism}). Instead, we introduce the premise of a trusted third-party service provider, which enjoys the trust of both user groups. Consequently, both Group 1 and Group 2 submit their raw data to this third-party service provider. This arrangement alleviates the need for direct data exchange between Group 1 and Group 2 while enabling the third-party to utilize the data for privacy mechanism training. It is important to emphasize that within our assumptions, the third-party service provider neither has access to any raw auxiliary dataset, nor has options to manually annotate certain portion of the data first. This assumption can be attributed to various reasons, including the absence of such a dataset, its non-public status, or the substantial cost and time required for its acquisition or annotation. Therefore, the provider is restricted to utilizing the data provided by the two user groups exclusively. Additionally, we assume that this third-party provider abstains from providing the user groups with trained privacy mechanism models. Instead, the third-party service provider exclusively provides black-box access i.e. it only delivers the resulting sanitized output data produced by the privacy mechanism model when given raw input data. This precautionary measure potentially mitigates extraction of data that was used to train the model. Hence, our framework operates akin to the principles observed in Machine Learning as a Service (MLaaS) applications, where users provide raw data and receive predictions, in our case, both user groups contribute raw data and receive sanitized data.

The raw data remains in a \textit{Restricted Access} block, accessible solely to a trusted third-party service provider until it undergoes sanitization. It is only when the privacy mechanism is trained and the raw data is sanitized that the data becomes available to the public or analysts. The sanitized dataset, denoted as $\hat{G}$, is depicted in Fig. \ref{problem_formulation_setup} in the \textit{Open Access} block. Therefore, for the training of models aimed at inferring private and utility features for Group 1, analysts can use the sanitized G2 data, referred to as $\hat{G}^2$. Likewise, analysts can infer Group 2's private and utility features by training models from $\hat{G}^1$ data.

\subsection{Philosophical Disagreements}
From a philosophical perspective, a realm of discourse emerges. Some individuals might assert that for specific groups of users, only the private features should differ while keeping the utility features constant, while others may propose the opposite, suggesting that only the utility features should vary while maintaining a consistent private feature. This choice is deeply rooted in philosophical contemplations and remains devoid of a universally definitive resolution. In our specific case, we assume the distinctiveness of all private and utility features. 

\subsection{Real World Application}
Our problem formulation has myriad potential real-world applications. Consider a scenario in which two hospitals possess datasets containing distinct sets of features. The first hospital seeks to predict a valuable feature that the second hospital possesses, while conversely, the second hospital aims to predict a different valuable feature possessed by the first hospital. Recognizing the costliness of data annotation or finding auxiliary datasets, both hospitals opt to mutually share their data. However, they soon realize that each hospital also hold labels for a feature that is private to them. Without privacy mechanisms, the risk arises that one hospital can infer the private features of the other's data during this exchange. Therefore, they opt for a collaborative approach where they refrain from direct data sharing and instead entrust their data to a trusted third-party service provider for sanitization and subsequent publication. Following data publication, both hospitals can leverage each other's data for predictive purposes without requiring additional data annotation, all while safeguarding the privacy of their respective private features. Likewise, there are numerous application domains where valuable inference privacy challenges can be effectively addressed by our setup.

\section{Methodology}\label{methodology}
In this section, we present our proposed methodology for sanitizing the dataset for two-group scenario. The primary objective of this methodology is to enable the precise prediction of utility features for both user groups while concurrently maintaining the privacy of these distinct groups. Our approach involves the introduction of a data-sharing mechanism that is capable of accommodating existing privacy mechanisms. First, we provide an overview of privacy mechanisms ALFR \cite{edwards2016censoring} and UAE-PUPET \cite{9892789} from existing literature, explaining their use in data sanitization for a single-group scenario. Later, we adapt these mechanisms for a two-group scenario.

\subsection{ALFR and UAE-PUPET}
\begin{figure}[!htb]
    \centering
    \includegraphics[width=0.455\textwidth]{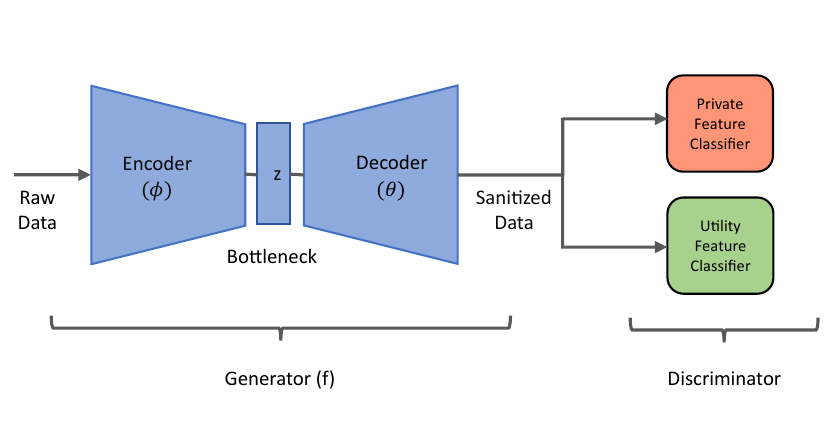}
    \caption{\textmd{ALFR and UAE-PUPET architecture}}
    \label{fig:architecture}
\end{figure}

ALFR (Adversarial Learned Fair Representations) \cite{edwards2016censoring} and UAE-PUPET (Uncertainty Autoencoder based Privacy and Utility Preserving end-to-end Transformation) \cite{9892789} are neural network architectures composed of two main components: a generator and a discriminator (refer to Fig. \ref{fig:architecture}). The generator comprises a generative model consisting of an encoder-decoder pair, while the discriminator includes classifiers for predicting private and utility features. Both ALFR and UAE-PUPET were designed to address privacy-utility tradeoffs in single-group scenarios.

In the context of a single group, it is assumed that a third-party service provider, responsible for training the privacy mechanism, has access to a raw auxiliary dataset (belonging to a virtual group $G3$) denoted as  $G^3$. The privacy mechanism (PM) is trained using data from the G3 dataset, and data from G1 is sanitized after the training of $PM^1$, where $G1$ represents the sole group in this single-group scenario, and $PM^1$ represents PM responsible for sanitizing G1 data.

\begin{algorithm}[!htb]
\small
\caption{ALFR and UAE-PUPET training algorithm}\label{alg:existing}
\begin{algorithmic}
\State $\gamma, \gamma_p, \gamma_u \gets$  initialize weights
\State $U \gets True$ 
\Repeat
\State $\hat{G}^3_{\mathcal{X}} \gets f(G^3_{\mathcal{X}})$,  $\hat{G}^3_{x^1_p} \gets s_p(\hat{G}^3_{\mathcal{X}})$, $\hat{G}^3_{x^1_u} \gets s_u(\hat{G}^3_{\mathcal{X}})$

\State $C \gets \text{MSE}(\hat{G}^3_{\mathcal{X}}, G^3_{\mathcal{X}})$ 

\State $l_p \gets \text{cross\_entropy}(\hat{G}^3_{x^1_p}, {G}^3_{x^1_p})$ 

\State $l_u \gets \text{cross\_entropy}(\hat{G}^3_{x^1_u}, {G}^3_{x^1_u})$ 

\State $L \gets \alpha C - \lambda_{p} l_p + \lambda_{u} l_u$ 
\If{ALFR}
\If {$U$}
\State Update $\gamma_p$ to \emph{maximize} the loss L
\Else
\State Update ($\gamma, \gamma_u$) to \emph{minimize} the loss L
\EndIf
\ElsIf{UAE-PUPET}
\If {$U$}
\State Update  $\gamma$ to \emph{minimize} the loss L
\Else
\State Update $\gamma_p$ to \emph{minimize} the loss $l_p$
\State Update  $\gamma_u$ to \emph{minimize} the loss $l_u$
\EndIf
\EndIf
\State $U \gets$ not $U$
\Until Deadline
\end{algorithmic}
\end{algorithm}

The generator, in its most general form, is represented as a function $f$, and its weights are denoted as $\gamma = (\phi, \theta)$. It takes the training input as $G^3_{\mathcal{X}}$, where the superscript denotes the group and the subscript denotes the feature(s), and generates sanitized training data $\hat{G}^3_{\mathcal{X}}$ as follows: $\hat{G}^3_{\mathcal{X}} = f(G^3_{\mathcal{X}})$. For the generator part, ALFR leverages a standard autoencoder (AE), while UAE-PUPET uses an Uncertainty Autoencoder (UAE) \cite{https://doi.org/10.48550/arxiv.1812.10539}. UAE introduces randomness and does not make any assumptions about Gaussian or any prior distribution on the latent variable or bottleneck, unlike the Variational Autoencoder \cite{kingma2022autoencoding}. The degree of randomness can be adjusted using a hyperparameter specified by the user. Both ALFR and UAE-PUPET utilize neural networks, denoted as functions $s_p$ and $s_u$, which act as simulated adversaries and utility provider networks and predict private and utility features represented  as $\hat{G}^3_{x^1_p}, \hat{G}^3_{x^1_u}$, respectively. The weights of $s_p, s_u$ are represented as $\gamma_p$ and $\gamma_u$, respectively. The simulated adversaries and utility provider's network contribute essential loss during the training stage.  The scalars $\alpha$, $\lambda_p$, and $\lambda_u$ are hyperparameters that determine the weighting of the competing objectives. This inclusion serves the purpose of introducing implicit regularization, involving noise introduction. This noise is intended to make the classification of private features challenging while allowing the classification of utility features. It is important to note that the introduced noise required for the task is solely influenced by the losses contributed by these discriminator models, and no additional noise is introduced. The optimizations within Algorithm \ref{alg:existing} provide a comprehensive view of the operational intricacies of ALFR and UAE-PUPET for the sake of completeness. In the upcoming section, we present our proposed \emph{data sharing mechanism} that extends the existing privacy techniques ALFR, and UAE-PUPET for managing privacy-utility tradeoffs in a two-group setting.

\subsection{Data Sharing Mechanism}\label{data-sharing-mechanism}
We can employ a unified adversarial architecture for training the privacy mechanism in a manner that facilitates data sanitization for both user groups. The unified architecture entails the integration of two additional classifiers into the discriminator for the second group's private and utility feature, coupled with the practice of alternating gradient updates between iterations by utilizing data from the alternate group. However, it is worth noting that adversarial privacy mechanisms are known to exhibit stability issues, particularly when the loss function lacks strong concavity \cite{pmlr-v130-xing21b}. Utilizing a single adversarial architecture while introducing additional optimizations for the second group may exacerbate these challenges. To confirm, we conducted experiments employing a unified architecture, and indeed, we observed the presence of stability issues. Therefore, we train separate adversarial architecture for each group, denoted as $PM^1$ and $PM^2$. In this setup, $PM^1$ is dedicated to the privatization of G1 data, while $PM^2$ is responsible for the privatization of G2 data.

The proposed data sharing mechanism follows an iterative and sequential process. Initially, the training of $(PM^1)_{m}$ for Group 1 is carried out, utilizing data from Group 2, specifically $G^2_{\mathcal{X}}$, along with labels $G^2_{x^1_p}$ and $G^2_{x^1_u}$. The subscript $m$ signifies the iterative process at time $m$, with $m=1$ as the starting point. During this training, the generator function $(f^1)_m$ is trained with $(s^1_p)_m$ and $(s^1_u)_m$. It is important to note that $(f^1)_m$ aims to learn how to sanitize data so that only the utility labels can be accurately inferred from the sanitized data. Once the privacy mechanism is trained, $(f^1)_m$ is utilitzed to generate sanitized data for Group 1, denoted as $(\hat{G^1_{\mathcal{X}}})_m$.

Considering that Group 1's data is now sanitized and intended for use by analysts in training predictive classifiers (if the sanitized data were to be published at this time), $(PM^2)_{m}$ chooses to leverage this sanitized data instead of the original Group 1 data for training its privacy mechanism catering to Group 2. Consequently, the generator function for $PM^2_{m}$, denoted as $(f^2)_m$, undergoes training using $(\hat{G}^1_{\mathcal{X}})_m$ data, in conjunction with labels $G^1_{x^2_p}$ and $G^1_{x^2_u}$, and utilizes $(s^2_p)_m$ and $(s^2_u)_m$ for $(PM^2)_{m}$.

\begin{figure}[!htb]
    \centering
    \includegraphics[width=0.45\textwidth]{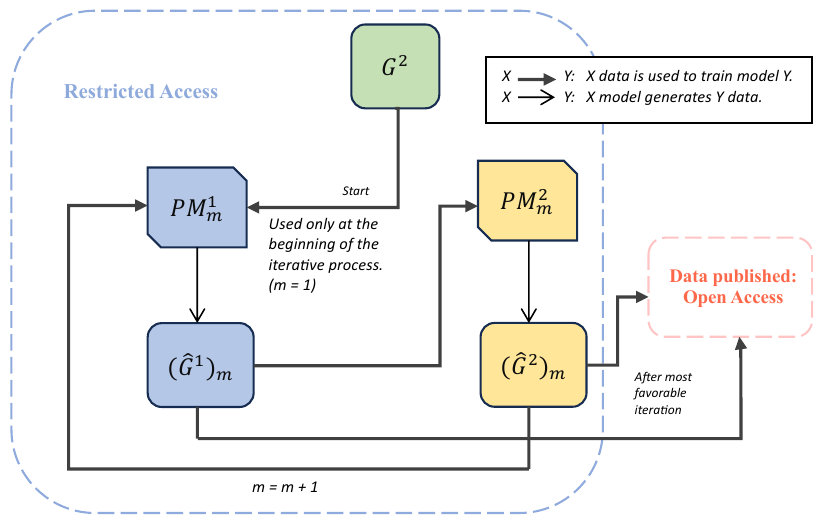}
    \caption{\textmd{Sequential and iterative nature of the data sharing approach. }}
    \label{iterative_process}
\end{figure}

After the training of the privacy mechanism, the generator function $(f^2)_m$ generates sanitized data for Group 2, referred to as $(\hat{G}^2_{\mathcal{X}})_m$. Subsequently, the third-party service provider, aware of the alterations in the data used to train the privacy mechanism $(PM^1)_{m}$, proceeds to retrain this privacy mechanism using the freshly sanitized data from Group 2. For Group 1, the third-party service provider now utilizes the updated sanitized data to retrain the privacy mechanism, denoted as $(PM^1)_{m+1}$, and Group 2 follows suit in retraining its privacy mechanism, referred to as $(PM^2)_{m+1}$, using the updated sanitized data from Group 1. In this context, the subscript $m+1$ signifies another round of the iterative process. This iterative process can be repeated for a total of $T$ rounds, where the index $m = \{1 \cdots T\}$.

The weights associated with the generators are meticulously preserved at each iteration $m$. The iteration that successfully achieves the desired privacy and utility guarantees is strategically utilized to generate the ultimate sanitized data for both groups, which is subsequently disclosed to the public. The final disclosed data from Group 1 is represented as $\hat{G}^1_{\mathcal{X}} \cup \{G^1_{x^2_p}, G^1_{x^2_u}\}$, while the ultimate disclosed data from Group 2 is denoted as $\hat{G}^2_{\mathcal{X}} \cup \{G^2_{x^1_p}, G^2_{x^1_u}\}$. Note that, the iterative process notation is removed once the process is completed. For a visual representation of the iterative process, please refer to Fig. \ref{iterative_process}.

We provide the algorithm for the data sharing mechanism in Algorithm \ref{alg:approach2}. This algorithm includes optimizations for a single iteration of a single group but can be iteratively applied to both groups using the mentioned sequence above or as visually demonstrated in Fig. \ref{iterative_process}. In this algorithm, the variable $i$ represents $PM^i$ being trained by Group $i$ using data from Group $j$. In the context of our scenario involving groups G1 and G2, if G1 is training $(PM^1)_m$, then $j = 2$ represents the data from G2 used for training. Similarly, if G2 is training $(PM^2)_m$, then $j = 1$. Furthermore, the subscript $m$ denotes the iteration number and primarily applies to the function weights. While we could have included the $m$ subscripts in the functions and losses as well, they are omitted here for readability purposes. For a detailed explanation of the notation used for various functions in the data sharing mechanism, please refer to Table \ref{tab:1}.

\begin{table}[!htb]
\scriptsize
  \centering
  \begin{tabular}{p{0.09\textwidth} | p{0.33\textwidth}}
  \toprule
\textbf{Notation} &  \textbf{Description} \\ 
 \midrule
$G^i_j$  & Raw data of Group $i$ with feature(s) $j$.\\
Example: $G^1_{\mathcal{X}}$ & Raw data of G1 i.e. rows of data that belong to G1 and have $\mathcal{X}$ features. \\
 \midrule
$\hat{G}^i_j$  & Sanitized data of Group $i$ with feature(s) $j$.\\
Example: $\hat{G}^2_{\mathcal{X}}$ & Sanitized data of G2 i.e. rows of data that belong to G2 and have $\mathcal{X}$ features. 
\newline \textit{Exception: If $\hat{G}^i_{x^i_p}$ / $\hat{G}^i_{x^i_u}$, it represents the private/utility feature predictions from $s^i_p / s^i_u$.} \\
 \midrule
$f^i$ & Function representing generator part of the privacy mechanism for Group $i$, ($PM^i$).\\
 \midrule
$s^i_p, s^i_u$ & Function representing simulated adversary and utility provider to predict the private and utility feature of Group $i$, respectively. \textit{Simulated} indicates adversaries and utilities network at the time of training. \\
 \midrule
$\gamma^i, \gamma^i_p, \gamma^i_u$ & Weights of $f^i, s^i_p, s^i_u$, respectively. \\
 \midrule
$l^i_p, l^i_u$ & Loss of a classifier attempting to infer the private and utility feature of Group $i$, respectively.\\
\bottomrule
  \end{tabular}
  \caption{\textmd{Essential notations and their descriptions.}}
  \label{tab:1}
\end{table}

\begin{algorithm}[!htb]
\small
\caption{Data Sharing Mechanism}\label{alg:approach2}
\begin{algorithmic}
\State $(\gamma^i)_m, (\gamma^i_{p})_m, (\gamma^i_{u})_m \gets$  initialize weights
\State $U \gets True$ 
\Repeat
\State $\hat{G}^j_{\mathcal{X}} \gets f^i(G^j_{\mathcal{X}})$, $\hat{G}^j_{x^i_p} \gets s^i_{p}(\hat{G}^j_{\mathcal{X}})$, $\hat{G}^j_{x^i_u} \gets s^i_{u}(\hat{G}^j_{\mathcal{X}})$

\State $C \gets \text{MSE}(\hat{G}^j_{\mathcal{X}}, G^j_{\mathcal{X}})$ 

\State $l^i_p \gets \text{cross\_entropy}(\hat{G}^j_{x^i_p}, {G}^j_{x^i_p})$ 

\State $l^i_u \gets \text{cross\_entropy}(\hat{G}^j_{x^i_u}, {G}^j_{x^i_u})$ 

\State $L \gets \alpha C - \lambda_{p} l^i_{p} + \lambda_{u} l^i_{u}$ 
\If {ALFR}
\If {$U$}
\State Update $(\gamma^i_{p})_m$ to \emph{maximize} the loss L
\Else
\State Update $(\gamma^i)_m,  (\gamma^i_{u})_m$ to \emph{minimize} the loss L
\EndIf

\ElsIf {UAE-PUPET}
\If {$U$}
\State Update $(\gamma^i)_m$ to \emph{minimize} the loss L
\Else
\State Update $(\gamma^i_{p})_m$ to \emph{minimize} the loss $l^i_p$
\State Update $(\gamma^i_{u})_m$ to \emph{minimize} the loss $l^i_u$
\EndIf
\EndIf
\State $U \gets$ not $U$
\Until Deadline
\end{algorithmic}
\end{algorithm}

\section{Experimental Setup}\label{experimental_setup}
\subsection{Dataset}
We conducted experiments using two distinct datasets: the US Census Demographic Data (referred to as \textit{US Census}) \cite{uscensus} and a synthetic dataset generated using the scikit-learn library \cite{scikit-learn}.  The US Census data is a 2017 American Community Survey dataset comprising 74,001 records, featuring 37 continuous variables. From the original 37 continuous variables, we selected 16 variables (e.g., male and female population, income, racial composition percentages, etc.), mirroring previous work \cite{9500410, 9892789}. Following preprocessing, the resulting dataset comprises 72,000 data points. The synthetic data used in our experiments also encompasses 16 continuous features and 72,000 data points. For both datasets, Groups G1 and G2 each consist of 31,000 distinct data points, while the remaining 10,000 data points are reserved as auxiliary datasets. Further details on the experiments involving auxiliary datasets are elucidated in Section \ref{auxiliarydataset}.

Our choice to utilitze the US Census data was guided by several factors. Firstly, our requirement was for a dataset comprising four categorical features, each appropriate for addressing balanced classification task, a type of dataset that is often rare and challenging to obtain. As a result, we selected this specific dataset, which initially comprises continuous features but can be transformed into balanced classification problems for all four attributes through discretization. These attributes represent the private and utility aspects of two distinct groups. Additionally, it is worth highlighting that this dataset has previously found application in research endeavors centered on exploring the balance between privacy and utility. Furthermore, our research maintains a specific focus on tabular datasets. However, the significant emphasis on continuous input features provides us with the potential to broaden our research horizons in future work, encompassing the incorporation of image datasets or datasets containing image embeddings. Importantly, this expansion can be achieved without the necessity for supplementary post-processing after data sanitizaion, a requirement typically associated when dealing with categorical input features \cite{9892789}.

\begin{figure*}[!htb]
\centering
\minipage{0.25\textwidth}
  \includegraphics[width = \textwidth]{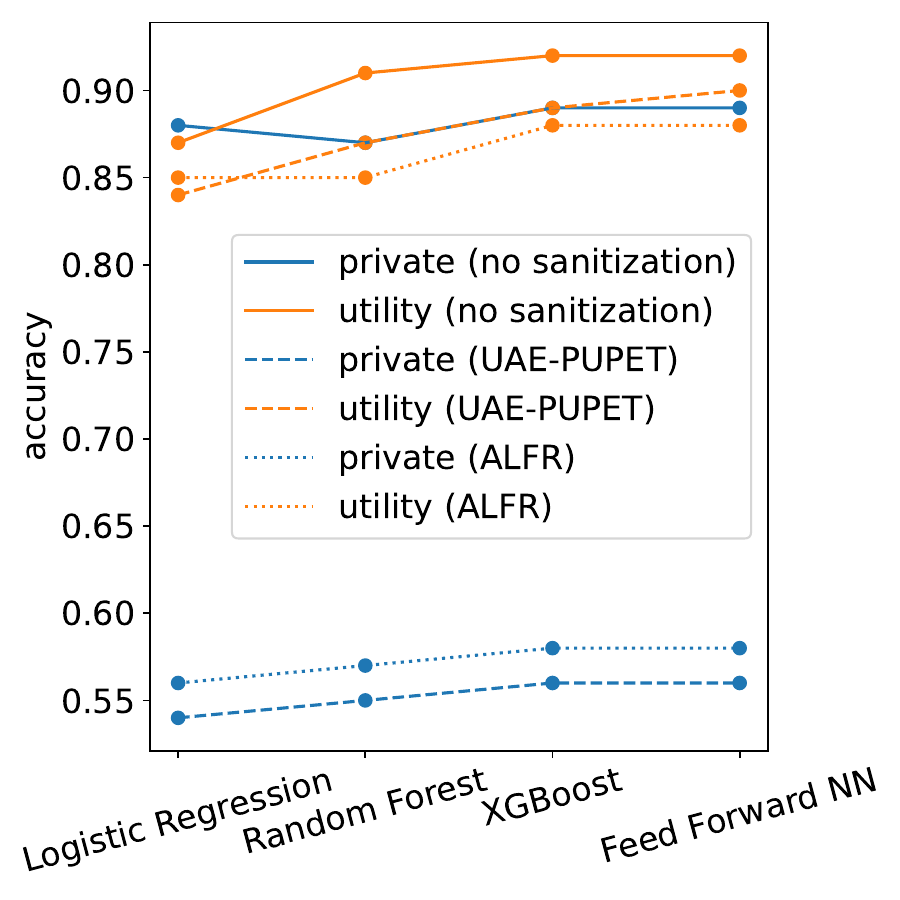}
  \subcaption{\scriptsize{US Census - G1}}
    \label{f1}
\endminipage
\minipage{0.25\textwidth}
  \includegraphics[width = \textwidth]{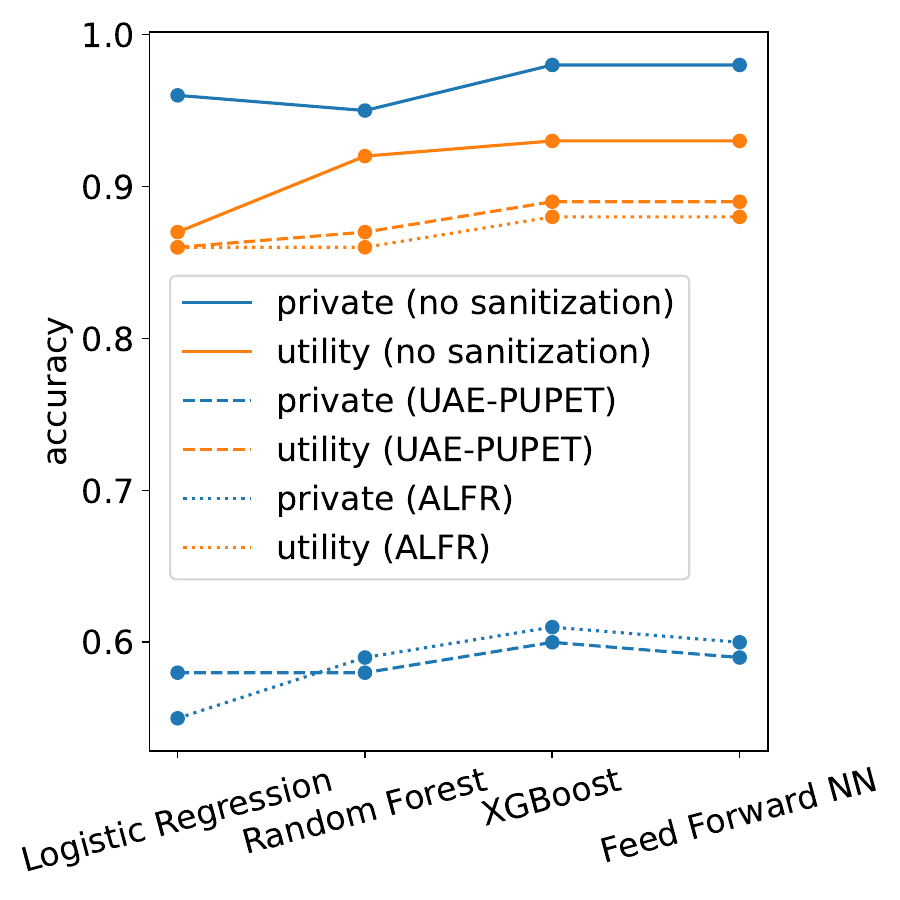}
  \subcaption{\scriptsize{US Census - G2}}
    \label{f2}
\endminipage
\minipage{0.25\textwidth}
  \includegraphics[width = \textwidth]{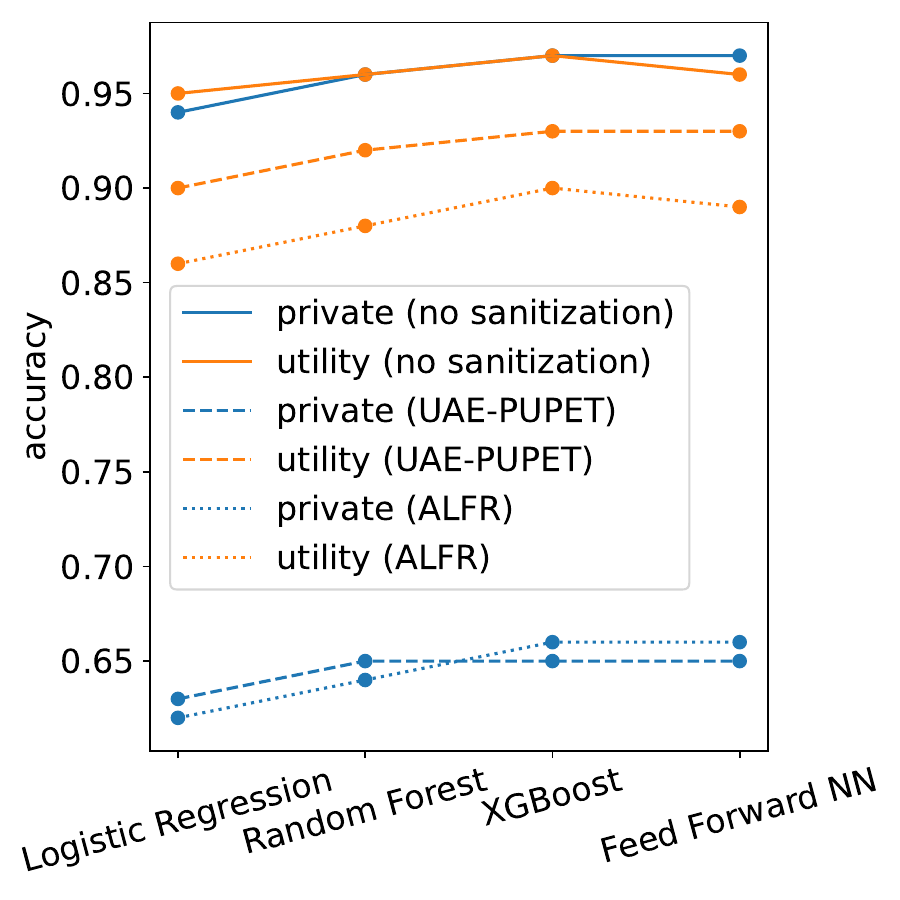}
  \subcaption{\scriptsize{Synthetic - G1}}
    \label{f3}
\endminipage
\minipage{0.25\textwidth}
  \includegraphics[width = \textwidth]{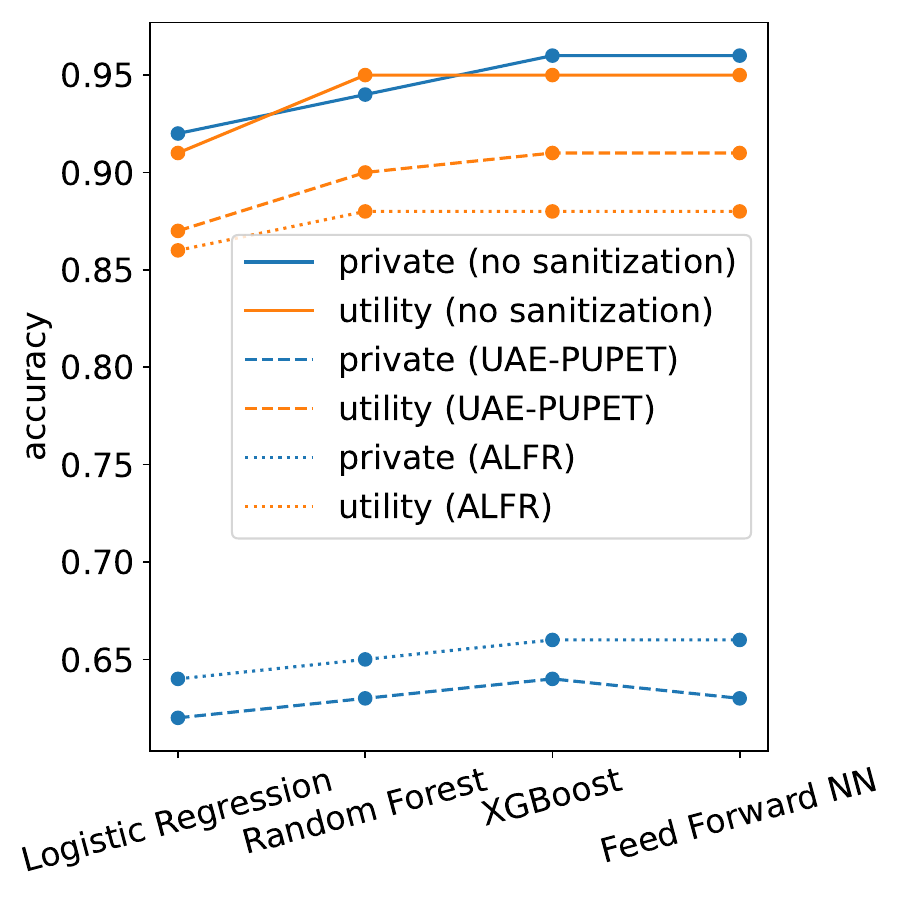}
  \subcaption{\scriptsize{Synthetic - G2}}
    \label{f4}
\endminipage
\caption{\textmd{Performance in predicting private and utility features, for both datasets and groups, before and after data sanitization. X-axis shows different ML models tested in this study, and Y-axis represents the accuracy of prediction.}}
\label{all_result}
\end{figure*}

\subsection{Private and Utility features}
Regarding the US Census dataset, the private feature for Group 1 (G1) refers to the variable \emph{Income}, whereas the of utility feature refers to the variable \emph{Employed} which refers to the number of employed population. This choice of private and utility feature is similar to the single-group setting in \cite{9500410, 9892789}. In our experimentation, we have categorized both private and utility attributes into two balanced groups each. For the Income variable, we have established categories based on whether the income exceeds 55000 or not. Correspondingly, for the Employed variable, values equal to or below 2000 are placed into one category, and values above this threshold constitute another category. Analogously, within Group 2 (G2), the private attribute is represented by the percentage of the white population, while the utility attribute is represented by the count of women in the population. For the white population percentage, we have delineated a category based on values surpassing 70, placing them into one class, and the rest into another class. Similarly, a class division has been established for the count of women, with values exceeding 2000 forming one category and those below constituting the other. These values have been thoughtfully selected after a thorough evaluation of the data distribution, ensuring a balanced classification scenario. The same methodology has been applied to synthetic data to attain a balanced classification setup.

\subsection{Performance Evaluation Metrics}
 To assess the performance of our privacy mechanism, we utilize standard accuracy and AUROC metrics. These metrics are expected to exhibit lower values when analysts attempt to predict private features, indicating stronger privacy protection, while for utility features, higher accuracy and AUROC values are desirable.

Additionally, we leverage empirical mutual information $(MI)$ \cite{gao2015efficient} to evaluate the information-theoretic correlation between our sanitized data and the private/utility features.

Furthermore, we utilize the privacy-utility tradeoff metrics introduced in \cite{10030960} where privacy leakage for attribute $p$ is defined as follows:
\begin{equation}
M_p = \frac{c_a(p) - c_r(p)}{c_n(p) - c_r(p)}
\label{eq:leakage1}
\end{equation}

Here, $c_n(.)$ denotes the accuracy of a classifier with raw, unsanitized data -- data without undergoing any privacy mechanism. Similarly, $c_a(.)$ represents the accuracy of a classifier after data sanitization, and $c_r(.)$ signifies the accuracy of the classifier when making random guesses (in our case, 0.5, as all are balanced binary classification problems). A lower $M_p$ value indicates closer proximity to a random guess and, consequently, stronger privacy guarantees.

Similarly, utility performance for attribute $u$ is quantified as follows:
\begin{equation}
M_u = \frac{c_a(u) - c_r(u)}{c_n(u) - c_r(u)}
\label{eq:gain1}
\end{equation}

In this case, a higher $M_u$ value signifies less utility drop. Finally, the privacy-utility tradeoff is calculated as:
\begin{equation}
T = \frac{M_p}{\delta + M_u}
\label{eq:tradeoff}
\end{equation}

Here, $\delta=0.0001$ is introduced to handle potential zero-division errors. A lower $T$ value indicates a more favorable tradeoff, implying that removing the privacy feature has a less negative impact on the utility feature.

\section{Result and Discussions}\label{resultsection}
In our study, we conduct 25 independent experiments to thoroughly evaluate the effectiveness of our data sharing mechanism. We report the mean and standard deviations  of the results to provide a comprehensive view of our approach's efficiency. However, in some instances, we omit reporting standard deviations for the sake of brevity, readability, or to enhance the clarity of figures. The aforementioned results correspond to the specific hyperparameter settings of $\alpha=1$, $\lambda_u=1$, and $\lambda_p=0.2$, which govern the relative weighting of the competing objectives related to privacy and utility attributes. 


\subsection{Accuracy and AUROC}
We utilize various ML models specifically Logistic Regression,  Random Forest, XGBoost, and Feed Forward Neural Network models to first train and predict private and utility features from unsanitized data. Subsequently, we employ proposed data-sharing mechanism utilizing both ALFR and UAE-PUPET techniques to sanitize the dataset. We then utilize the sanitized data to retrain the machine learning models from scratch and predict private and utility features. The outcomes of these diverse machine learning models on unsanitized and sanitized data are summarized in Fig. \ref{all_result}, where each plot represents a distinct dataset and group. For example, Fig. \ref{f1} presents results for Group 1 (G1) of the US Census dataset. The X-axis in these plots represents the various models used, while the Y-axis signifies the corresponding accuracy achieved by each machine learning model. Note that, each point on the plots represent mean accuracy scores of 25 experiments. Overall, for our dataset, these plots indicate that XGBoost and Feed Forward Neural Networks consistently attain the highest accuracy among the models considered for classification.

\begin{table}[!htb]
\centering
\scriptsize
\begin{tabular}{l c c c c}
\toprule
\multicolumn{1}{c}{\multirow{2}{*}{\begin{tabular}[c]{@{}c@{}}\textbf{Privacy} \\ \textbf{Mechanism}\end{tabular}}} & \multicolumn{2}{c}{\textbf{Group 1 (G1)}} & \multicolumn{2}{c}{\textbf{Group 2 (G2)}} \\  \cmidrule{2-5} 
\multicolumn{1}{c}{}                                                                              & \textbf{Private}         & \textbf{Utility}        & \textbf{Private}         & \textbf{Utility}        \\ \midrule
 \multicolumn{5}{l}{\textit{US Census Dataset }}                                                                                                            \\ \midrule
no privacy     & 0.88 $\pm$ 0.002    & 0.92   $\pm$ 0.003  & 0.98 $\pm$ 0.004     & 0.93   $\pm$ 0.006       \\ 
alfr    & 0.58 $\pm$ 0.005        & 0.87  $\pm$ 0.014          & 0.60  $\pm$ 0.011          & 0.88  $\pm$ 0.011         \\ 
uae-pupet    & 0.55  $\pm$ 0.003          & 0.90  $\pm$ 0.004          & 0.59  $\pm$ 0.012          &  0.89     $\pm$ 0.009      \\ \midrule

\multicolumn{5}{l}{\textit{Synthetic Dataset}}                                                                                                               \\ \midrule
no privacy     & 0.97 $\pm$ 0.003           & 0.96  $\pm$ 0.008          & 0.97  $\pm$ 0.003          & 0.95    $\pm$ 0.004       \\ 
alfr     & 0.66  $\pm$ 0.007         & 0.90  $\pm$ 0.009         & 0.66  $\pm$ 0.004           &  0.89  $\pm$ 0.012        \\ 
uae-pupet       & 0.65  $\pm$ 0.005          & 0.92 $\pm$ 0.008          & 0.63 $\pm$ 0.006           &  0.92   $\pm$ 0.008           \\ \bottomrule
\end{tabular}
\caption{\textmd{Accuracy scores of predicting private and utility features before and after employing the proposed data-sharing mechanism using existing  ALFR (alfr) and UAE-PUPET(uae-pupet) techniques.}}
\label{tab:results1}
\vspace{-0.075in}
\end{table}

Among the four models employed, we identify the highest accuracy achieved as the benchmark accuarcy for predicting private or utility attributes within a particular group. We summarize these accuracy results in Table \ref{tab:results1}. Notably, the accuracy and AUROC scores are nearly identical, so we omit the AUROC scores for brevity. From the table, we can see that for G1 in the US Census dataset without any data sanitization (no privacy), the accuracy is 0.88 for private features and 0.92 for utility features. However, after applying the data sharing mechanism using ALFR, the accuracy for private features drops to 0.58, and it drops to 0.55 when using the UAE-PUPET technique. In contrast, the accuracy for utility features remains high at 0.87 with ALFR and 0.90 with UAE-PUPET.

Similarly, for G2, the accuracy for private features decreases from 0.98 to 0.60 with ALFR and to 0.59 with UAE-PUPET, while the accuracy for utility features experiences a slight decrease from 0.93 to 0.88 with ALFR and 0.89 with UAE-PUPET. For results related to the synthetic dataset, please refer to Table \ref{tab:results1}. Overall, our proposed data sharing mechanism appears to effectively safeguard both privacy and utility. Specifically, the use of UAE-PUPET in the data-sharing mechanism consistently yields lower accuracy for private features and better accuracy for utility features compared to ALFR.

\subsection{Mutual Information}

\begin{figure}[!htb]

\centering
\minipage{0.245\textwidth}
  \includegraphics[width = \textwidth]{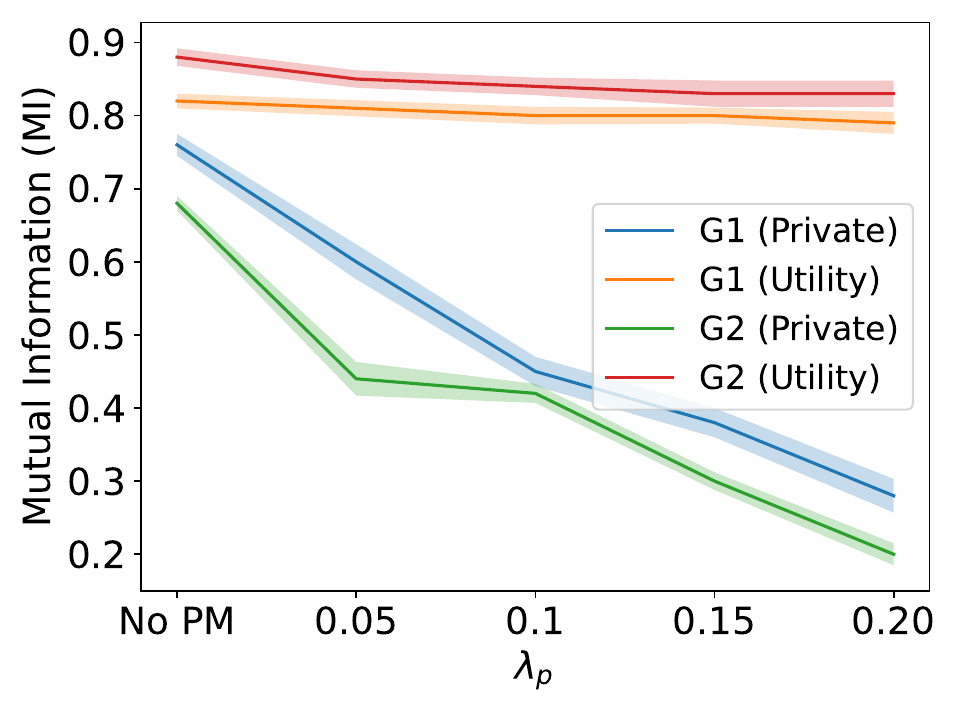}
  \subcaption{\scriptsize{US Census Dataset}}
    \label{mi1}
\endminipage
\minipage{0.245\textwidth}
  \includegraphics[width = \textwidth]{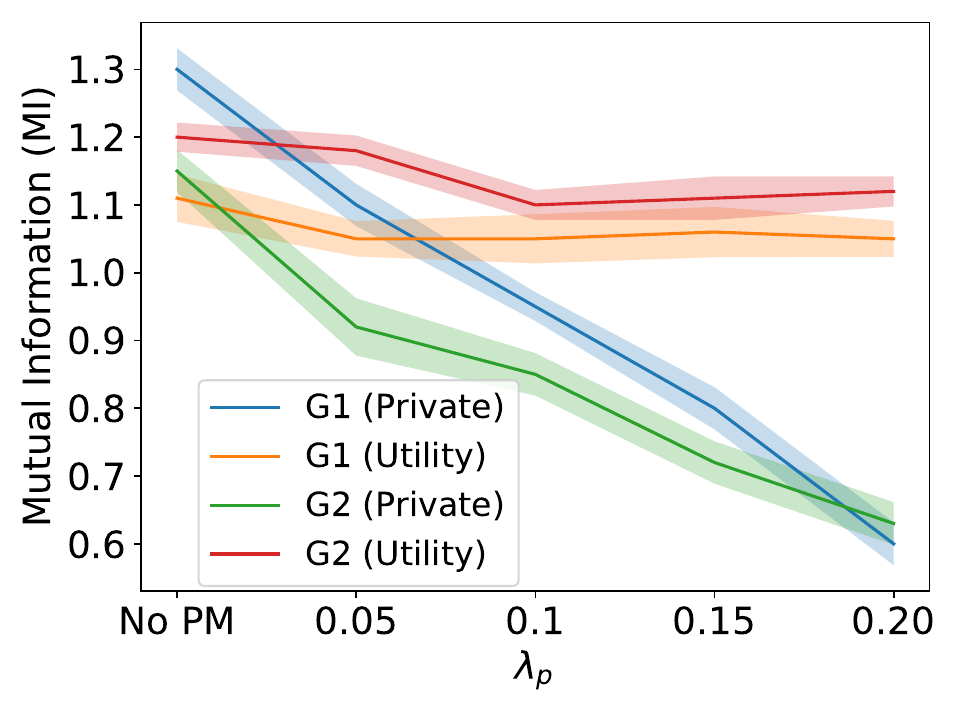}
  \subcaption{\scriptsize{Synthetic Dataset}}
    \label{mi2}
\endminipage
\caption{\textmd{Mutual Information (MI) between sanitized data derived with various $\lambda_p$ values and private/utility labels. \textit{No PM} on the X-axis indicates MI between the original(unsanitized) data and private/utility labels.
} }
\label{mutualinformation}
\vspace{-0.07in}
\end{figure}

We employ mutual information as an information-theoretic measure to evaluate whether our sanitized data exhibits reduced correlation with private labels while preserving similar correlation with utility features. Fig. \ref{mutualinformation} visually demonstrates the efficacy of our data-sharing mechanism in diminishing the mutual information between the sanitized data and private features while simultaneously maintaining the mutual information between the sanitized data and utility features. The parameter $\lambda_p$ signifies the weighted objective assigned to the privacy mechanism's loss function. A higher $\lambda_p$ value, while keeping $\alpha$ and $\lambda_u$ constant, indicates a heightened level of privacy protection. The mutual information (MI) values depicted in the figure reflect the outcomes achieved after the incorporation of UAE-PUPET's technique within our data-sharing mechanism.

\begin{table}[!htb]
\centering
\scriptsize
\begin{tabular}{ccccc}
\toprule
\multicolumn{1}{c}{\multirow{2}{*}{\begin{tabular}[c]{@{}c@{}}Performance \\ Metrics\end{tabular}}} & \multicolumn{2}{c}{US Census Dataset} & \multicolumn{2}{c}{Synthetic Dataset} \\ \cmidrule{2-5} 
\multicolumn{1}{c}{}      & Group 1       & Group 2       & Group 1     & Group 2       \\ \midrule
$M_p$      &  0.20, \textbf{0.15}        &  0.22, \textbf{0.20}        &  0.34, \textbf{0.31}      &  0.34, \textbf{0.30}      \\ 
$M_u$        &  0.90, \textbf{0.95}       &  0.88, \textbf{0.90}   &  0.85, \textbf{0.91}      &  0.84, \textbf{0.91}      \\ 
$T$        &  0.22, \textbf{0.15}       &  0.25, \textbf{0.22}      &  0.40, \textbf{0.34}      &  0.40, \textbf{0.32}      \\ \bottomrule
\end{tabular}
\caption{\textmd{Privacy Leakage, Utility Performance, and Privacy Utility Tradeoffs. Bold highlights the superior method between ALFR and UAE-PUPET.}}
\label{tab:results2}
\end{table}

\begin{figure*}[!htb]
\centering
\minipage{0.22\textwidth}
  \includegraphics[width = \textwidth]{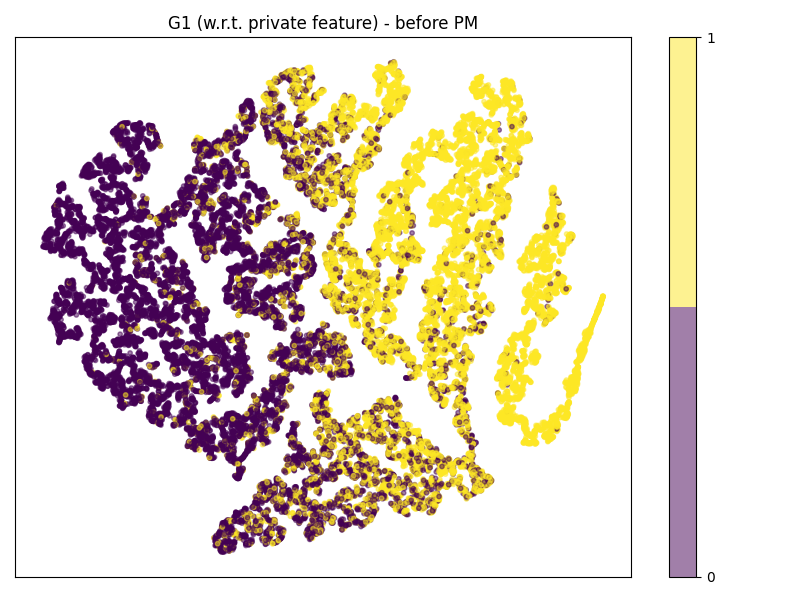}
  \subcaption{\scriptsize{G1-Private (before)}}
    \label{g1-p-before}
\endminipage
\minipage{0.22\textwidth}
  \includegraphics[width = \textwidth]{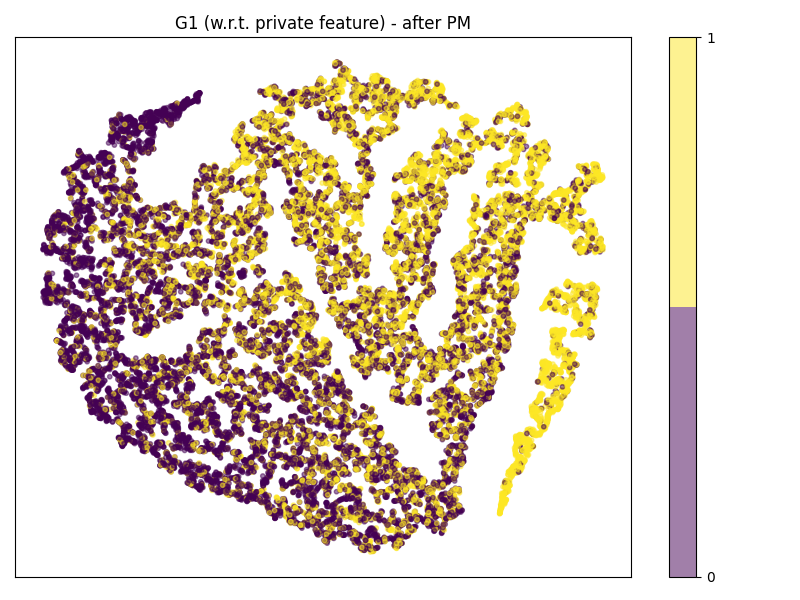}
  \subcaption{\scriptsize{G1-Private (after)}}
    \label{g1-p-after}
\endminipage
\minipage{0.22\textwidth}
  \includegraphics[width = \textwidth]{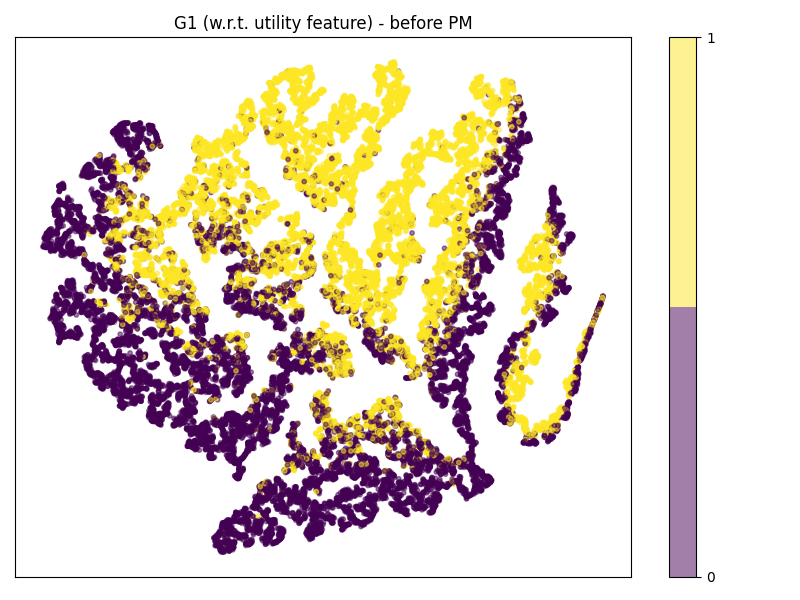}
  \subcaption{\scriptsize{G1-Utility (before)}}
    \label{g1-u-before}
\endminipage
\minipage{0.22\textwidth}
  \includegraphics[width = \textwidth]{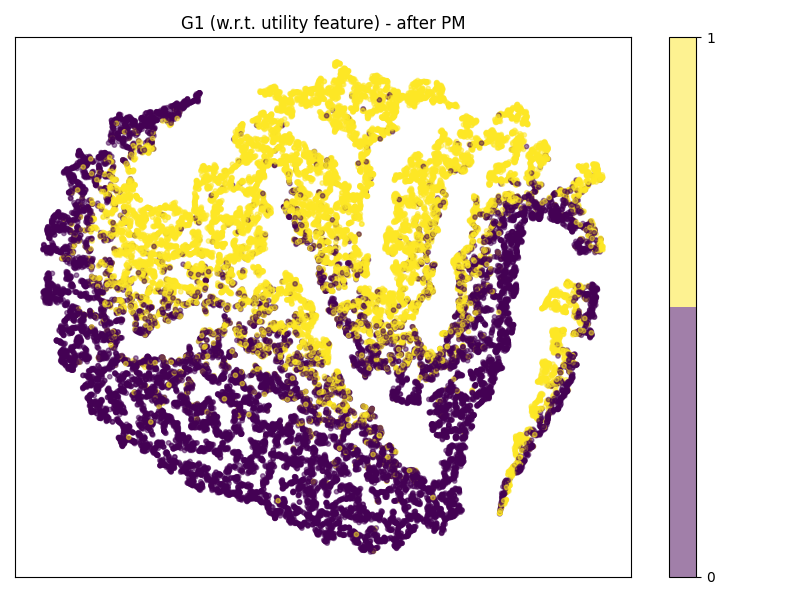}
  \subcaption{\scriptsize{G1-Utility(after)}}
    \label{g1-u-after}
\endminipage

\minipage{0.22\textwidth}
  \includegraphics[width = \textwidth]{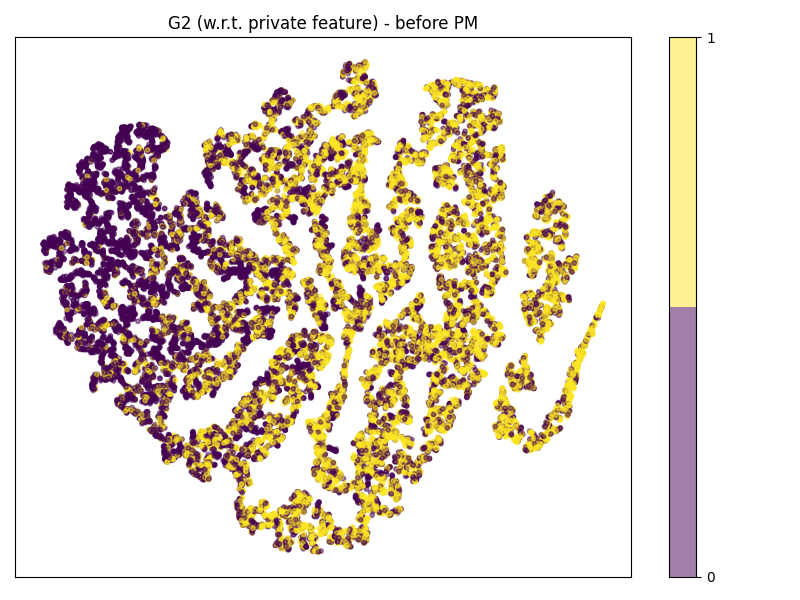}
  \subcaption{\scriptsize{G2-Private (before)}}
    \label{g2-p-before}
\endminipage
\minipage{0.22\textwidth}
  \includegraphics[width = \textwidth]{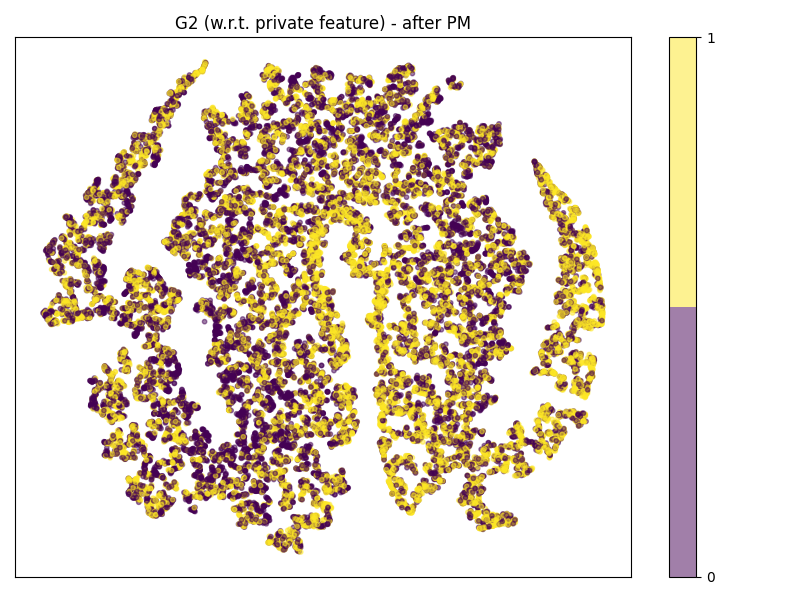}
  \subcaption{\scriptsize{G2-Private (after)}}
    \label{g2-p-after}
\endminipage
\minipage{0.22\textwidth}
  \includegraphics[width = \textwidth]{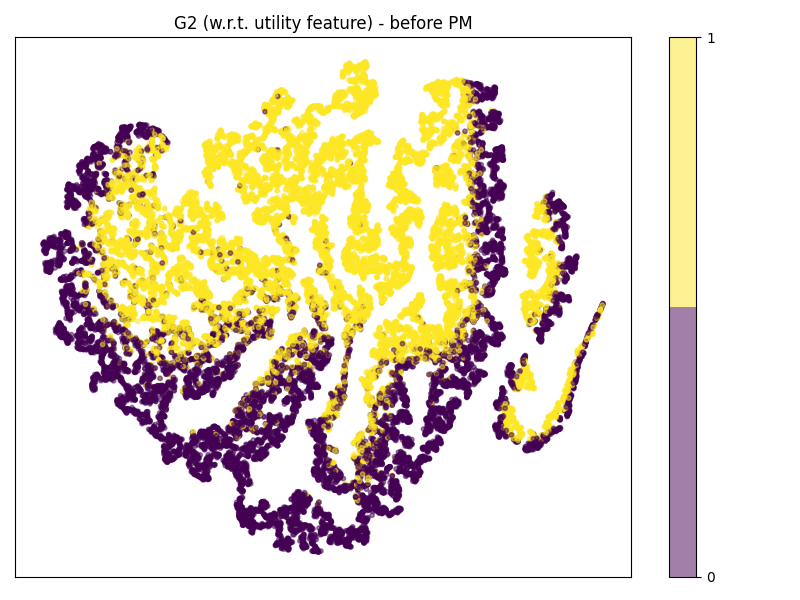}
  \subcaption{\scriptsize{G2-Utility (before)}}
    \label{g2-u-before}
\endminipage
\minipage{0.22\textwidth}
  \includegraphics[width = \textwidth]{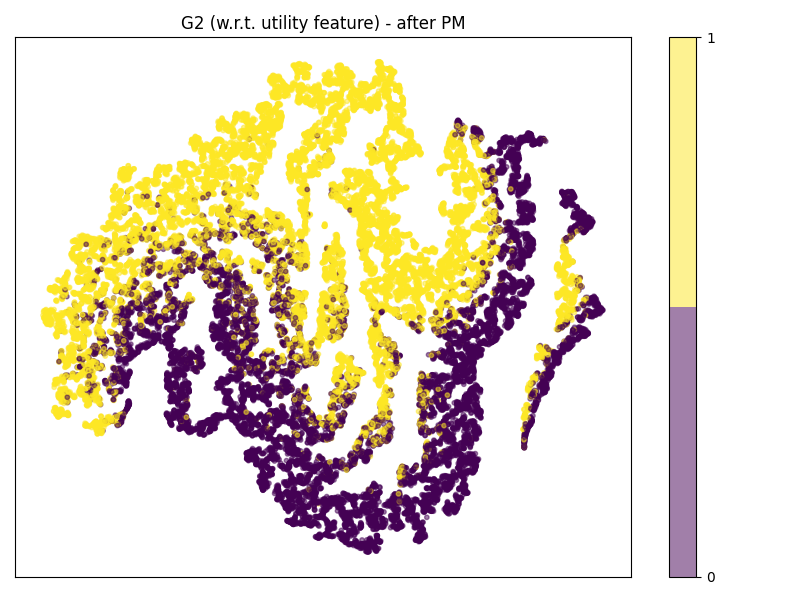}
  \subcaption{\scriptsize{G2-Utility (after)}}
    \label{g2-u-after}
\endminipage

\caption{\textmd{Visualization of unsanitized and sanitized data w.r.t private and utility labels. \textit{before} represents unsanitized data whereas, \textit{after} represents sanitized data using UAE-PUPET technique.} }
\label{fig-last}
\vspace{-0.075in}
\end{figure*}

\subsection{Privacy, Utility and Tradeoffs}

Table \ref{tab:results2} illustrates the result of the following metrics: Privacy Leakage ($M_p$), Utility Performance ($M_u$), and Privacy-Utility Tradeoff ($T$). The values in the form of \textit{a}, \textit{b} correspond to results obtained through the ALFR and UAE-PUPET techniques, respectively. For instance, in the US Census Dataset for Group 1, $T$ is 0.22 (ALFR) and 0.15 (UAE-PUPET).  Lower $T$ value indicates a more favorable tradeoff  implying that removing the privacy feature has a less negative impact on the utility feature. Overall, the table signifies the effectiveness of the data-sharing mechanism with both ALFR and UAE-PUPET techniques. In addition, the table highlights UAE-PUPET consistently outperforms ALFR within the data-sharing mechanism. However, the primary aim of this paper is not to determine the ultimate optimal approach. Rather, it serves to introduce an innovative and real world problem formulation. Moreover, the paper seeks to foster increased involvement from researchers within this field, with the aspiration that our findings can establish a standardized benchmark for forthcoming investigations.

\subsection{Data Visualization in two-dimensions}
We employ t-SNE (t-distributed Stochastic Neighbor Embedding) to create two-dimensional visualizations of both the unsanitized and sanitized US Census data concerning private and utility labels, as shown in Fig. \ref{fig-last}. The visualizations reveal noteworthy patterns. Before and after sanitization, the utility feature exhibits a clear separation into distinct clusters for the positive and negative classes. However, after sanitization, the private feature lacks such distinct clusters for the positive and negative categories. 


\subsection{Is privacy affected if analyst collects an auxiliary dataset?}\label{auxiliarydataset}
We further explore scenarios in which an analyst possesses the capacity to acquire an additional dataset for the purpose of training various machine learning models to predict private features. Our aim is to evaluate whether the incorporation of this auxiliary dataset ($G^3$ data) can enhance the analyst's predictive performance. The analyst, who can also be considered an adversary, employs two distinct strategies. The first strategy involves training exclusively on the auxiliary dataset, comprising 10,000 data points exclusive to those in Group 1 and Group 2. The second strategy entails training using a combination of both the auxiliary dataset and a sanitized dataset. It is observed that both strategies yield similar predictive accuracies for the private feature, as depicted in Table \ref{tab:results1}, and none of the strategies demonstrate higher accuracy than what is presented in Table \ref{tab:results1}. This reinforces our assertion that our data sharing mechanism effectively safeguards the privacy of private features for both user groups, regardless of whether the analyst only has access to the sanitized data or can acquire an auxiliary dataset for training various predictive models. 

\section{Conclusion and Future Work}\label{conclusion}
In this paper, we introduce a novel problem formulation that focuses on balancing the tradeoff between privacy and utility within two distinct user groups. Notably, the third-party responsible for sanitizing the dataset has no access to any auxiliary dataset and is not required to annotate data separately for different user groups. Instead, the proposed data sharing mechanism leverages alternate group data. Additionally, analysts are restricted to utilizing only the sanitized data for training their predictive models. Our proposed data sharing mechanism can be seamlessly integrated with existing privacy techniques. In our work, we demonstrate the applicability of the data sharing mechanism in conjunction with ALFR and UAE-PUPET. Both of these techniques, when incorporated into the data sharing mechanism, have proven effective in reducing the analyst's accuracy in predicting private features while maintaining a high level of accuracy in predicting utility features.
Furthermore, our experiments indicate that even when analysts gather auxiliary datasets, the privacy of both user groups remains intact. Although our current research is focused on a two-group setting using tabular datasets, we plan to extend our investigations to scenarios involving an arbitrary number of groups, encompassing both tabular and image datasets.

The primary objective of this paper is to introduce a novel real-world problem formulation, addressing a gap that exists despite the substantial research efforts dedicated to the intricate landscape of the privacy-utility tradeoff. Moreover, the problem formulation and the proposed solution have the potential for broader applications in mitigating biases, enhancing fairness, and rectifying discrimination in machine learning models, especially in scenarios involving multiple groups. Additionally, our aim is to inspire more researchers in this field to engage with this novel problem, with the hope that our findings can serve as a benchmark for future research endeavors.

\bibliographystyle{IEEEtran}
\bibliography{IEEEabrv, references}

\end{document}